\documentclass[letterpaper, 10 pt, conference]{ieeeconf}  

\IEEEoverridecommandlockouts                              

\usepackage{times}
\usepackage{epsfig}
\usepackage{graphicx}
\usepackage{amsmath}
\usepackage{amssymb}
\usepackage{epstopdf}
\usepackage{amsfonts,amssymb} 
\usepackage{amsmath, bm}
\usepackage{float}
\usepackage{tabu}
\usepackage{multirow}
\usepackage[table,xcdraw]{xcolor}
\usepackage{longtable}
\usepackage{booktabs}
\usepackage{setspace}
\usepackage{pifont}
\usepackage{tablefootnote}
\usepackage{color}

\makeatletter
\let\NAT@parse\undefined
\makeatother
\usepackage[pagebackref=true,breaklinks=true,letterpaper=true,colorlinks,bookmarks=false]{hyperref}
\usepackage{url}
\usepackage{booktabs}
\usepackage{multirow}
\def\eg{\textit{e.g.}}
\def\etal{\textit{et al.}}

\def\ie{\textit{i.e.}}

\newcommand{\revise}[1]{\textcolor[rgb]{0,0,0}{#1}}
\title{\LARGE \bf
I Know What You Draw: \\Learning Grasp Detection Conditioned on a Few Freehand Sketches 
}

\author{Haitao Lin$^{1}$, Chilam Cheang$^{1}$, Yanwei Fu$^{1}$ and Xiangyang Xue$^{1}$
\thanks{*This work was supported in part by NSFC under Grant (No. 62076067), STCSM Project (19511120700), and Shanghai Municipal Science and Technology Major Project (No.2021SHZDZX0103).}%
\thanks{$^{1}$Fudan University. \{htlin19,ccheang19,yanweifu,xyxue\}@fudan.edu.cn. Yanwei Fu is the corresponding author, School of Data Science.}%

}

\begin{document}

\maketitle
\thispagestyle{empty}
\pagestyle{empty}

\begin{abstract}
In this paper, we are interested in the problem of generating target grasps by understanding freehand sketches. The sketch is useful for the persons who cannot formulate language and the cases where a textual description is not available on the fly. However, very few works are aware of the usability of this novel interactive way between humans and robots. To this end, we propose a method to generate a potential grasp configuration relevant to the sketch-depicted objects. Due to the inherent ambiguity of sketches with abstract details, we take the advantage of the graph by incorporating the structure of the sketch to enhance the representation ability. This graph-represented sketch is further validated to improve the generalization of the network, capable of learning the sketch-queried grasp detection by using a small collection (around 100 samples) of hand-drawn sketches. Additionally, our model is trained and tested in an end-to-end manner which is easy to be implemented in real-world applications. Experiments on the multi-object VMRD and GraspNet-1Billion datasets demonstrate the good generalization of the proposed method. The physical robot experiments confirm the utility of our method in object-cluttered scenes \footnote{Project webpage. \url{https://hetolin.github.io/Skt_grasp}}.

\end{abstract}

\section{INTRODUCTION}

\begin{figure*}[thpb]
  \centering
  \includegraphics[width=0.85\textwidth]{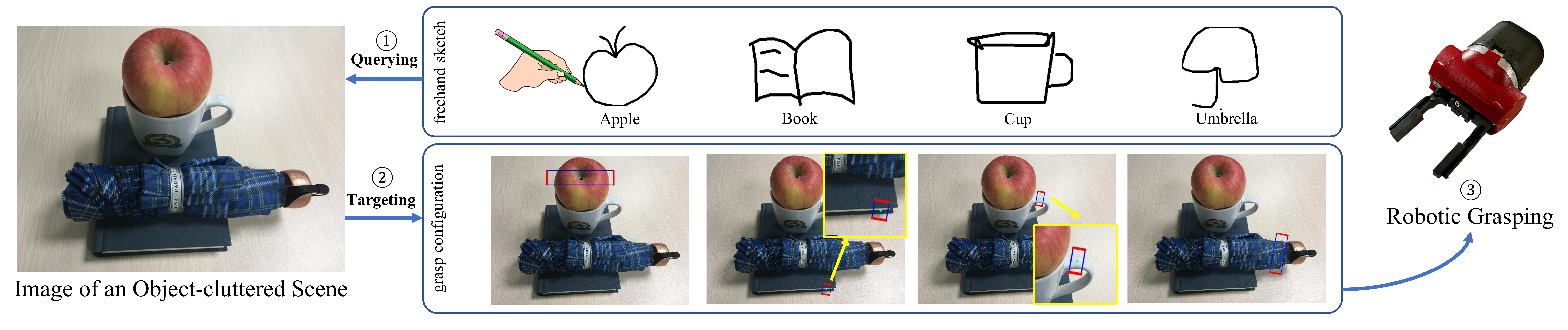}
  \vspace{-0.1in}
  \caption{An overview of our task. Given a natural image of an object-cluttered scene, we utilize various freehand sketches to query the potential grasp configurations in this natural image. The robot grasps target objects depicted by the sketch according to the predicted grasp configuration.\label{fig:overview}}
  \vspace{-0.15in}
\end{figure*}

Napoleon Bonaparte wrote that 
`\textit{A good sketch is better than a long speech}'.
Compared to human language, sketches are a more intuitive way for humans to describe visual objects around the world. Recently, many great efforts are made to enable the robot to understand the languages~\cite{chen2021joint, guadarrama2014open, nguyen2020robot, rao2018learning, hatori2018interactively, shridhar2018interactive}. In contrast,
it has been less touched on the `sketch-driven' human-robot interaction, which however is very important to some real-world applications beyond linguistic descriptions.
For example, the robot capable of understanding sketches should have the ability to communicate with aphasiac who cannot formulate the sentences~\cite{damasio1992aphasia}, play with children who cannot speak fluently yet, or keep the autistic children accompanied. In smart home applications, the mobility-impaired elders may communicate with the robots by sketching to grab something they happen to forget the name. 
This motivates our task of grasping target objects from the freehand sketch query.

Typically, it is very nontrivial to learn to grasp objects by freehand sketches. We highlight there exist several challenges: (1) The freehand sketch is a sparse representation expressing the objects in abstract, descriptive, or implied level, (2) the sketches have large \textit{intra-class variations}, as the style of each sketch is unique from each individual,
(3) There are \textit{large inter-domain variations} between the sketch and natural image; this reflects the different distribution of raw pixels, and (4) it is very difficult to obtain a large amount of paired sketches and images to train the models. As a result, such models have to be trained by a few sketches.

To address these problems, we present the Graph Convolutional Network (GCN) learning to extract the representations of sketches. 
Our GCN learning is built upon the graph structure of sketches, and thus efficiently reduces the ambiguity and intra-class variations of sketches. Critically, such graph representations of sketch help accelerate the converge of the network and improve the performance with only a few-shot sketch input. To reduce the domain gap,
 we thus introduce a network to optimize the relevance between paired sketches and images, where the sketch features from GCN are utilized as queries to search relevant grasp proposals using the features computed on images.

In this paper, we present a sketch-conditioned grasp retrieval method by detecting potential grasps of target objects depicted by freehand sketches as shown in Fig.~\ref{fig:overview}. This method takes an RGB image and a freehand sketch as input. 
\revise{The pixels of the sketch are further sub-sampled and selected as nodes to construct a graph representation, which incorporates information of the adjacent points in each stroke.}
\revise{Compared to pixel-based sketch, graph-based sketch} helps the network extract more representative features to differentiate the sketch, facilitating the generalization ability of the network. Therefore, \revise{encoding structural information of the sketch based on GCN is necessary, which} enables our pipeline to be efficiently trained with \textit{a few sketch samples} (around 100 samples).
We perform a feature-level query by using the feature maps of the natural image and corresponding sketch to obtain image-sketch relevance feature maps.
Then the relevance feature maps are adopted by the grasp proposal network for retrieving the grasp region candidates relevant to the sketch. Finally, the grasp detection network treats the grasp orientation learning as a multi-class classification task and regresses the 5-dimensional grasp parameters based on the grasp proposals.

\noindent \textbf{Contributions.} This paper presents three core technical contributions: 
1) We study a novel human-robot interaction way that enables understanding the content of sketches on the robots for conditional grasping.
2) We design an end-to-end network for learning sketch-queried grasp detection for robots.
3) We represent the sketch as a graph, and this helps our network have good generalization ability in the few-shot task by using a few sketches.
We evaluate our method on the object-cluttered VMRD~\cite{zhang2018visual} and GraspNet-1Billion~\cite{fang2020graspnet} datasets. We also conduct physical robot experiments on the Baxter robot with a collection of 24 household objects (Fig.~\ref{fig:objects}) to validate the efficacy of our proposed method.

\section{Related Work}
\noindent \textbf{Robot Understanding the Sketch.}
 Most recent works~\cite{chen2021joint, guadarrama2014open, nguyen2020robot, rao2018learning, hatori2018interactively, shridhar2018interactive} devote to developing the ability to understand verbal instructions on the robot for providing intuitive human-robot interaction. Nevertheless, freehand sketch provides a more convenient way for humans to communicate with robots, such as depicting indescribable objects, which do not have verbal instructions. The hand-drawn sketch is less explored in service robots. 
 \revise{Some recent works have employed sketches as an easier way to navigate the robot than scanned metrical maps~\cite{shah2010robust, skubic2002hand,boniardi2015robot,boniardi2016autonomous, ahmad2021sketch}.}
 To the best of our knowledge, there is no previous work capable of grasping objects associated with sketch content.

\noindent \textbf{Grasp Detection.} It detects all potential graspable regions around the objects, where grasp configuration is often represented as an oriented grasping rectangle proposed in~\cite{jiang2011efficient, lenz2015deep}. Some works~\cite{lenz2015deep, jiang2011efficient, depierre2018jacquard, guo2017hybrid, redmon2015real, kumra2017robotic} focus on detecting a single grasp on the single-object scenes. Furthermore,  methods~\cite{chu2018real, zhou2018fully, asif2019densely, morrison2018closing} show the ability of grasp detection in more cluttered scenes including multiple objects.
Specially, with the success paradigm of Region Proposal Network (RPN)~\cite{ren2015faster}, works of~\cite{zhang2019roi,chu2018real} have shown promising performance by exploring such deep object detection network to solve grasp detect problem. We also incorporate the design principle of RPN and further propose the end-to-end network to solve conditional grasp detection by grounding the depicted object to the abstract sketch.

\noindent \textbf{Sketch-based Image Retrieval (SBIR).} It aims to query natural images associated with the given freehand sketch. 
Traditional methods usually retrieve by leveraging the hand-crafted descriptors~\cite{lowe2004distinctive,hu2010gradient,hu2011bag} to extract features.
Recent deep neural networks have gained promising performance over the traditional methods due to their strong representation ability and capacity. Various kinds of networks and loss functions~\cite{qi2016sketch, sangkloy2016sketchy, yu2016sketch,song2017deep} are designed to solve large domain gap between sketch queries and natural images. 
In contrast to these works, we focus on directly \textit{detecting the potential grasp configuration in the cluttered scene images}. Our task is more challenging than simply retrieving images with salient objects: given the input sketches, it has to search target grasp configurations in object-cluttered scenes. 
\revise{The most relevant approach to ours is the work of Tripathi~\etal~\cite{eitz2010sketch}. The critical difference is that our grasp-level localization is represented by an oriented rectangle, which is helpful for robotic grasping in object-clutter scenes. However, theirs is object-level localization represented by an axis-aligned rectangle in scenes with less occlusion between objects.}


\begin{figure}[thpb]
  \centering
  \includegraphics[scale=0.3]{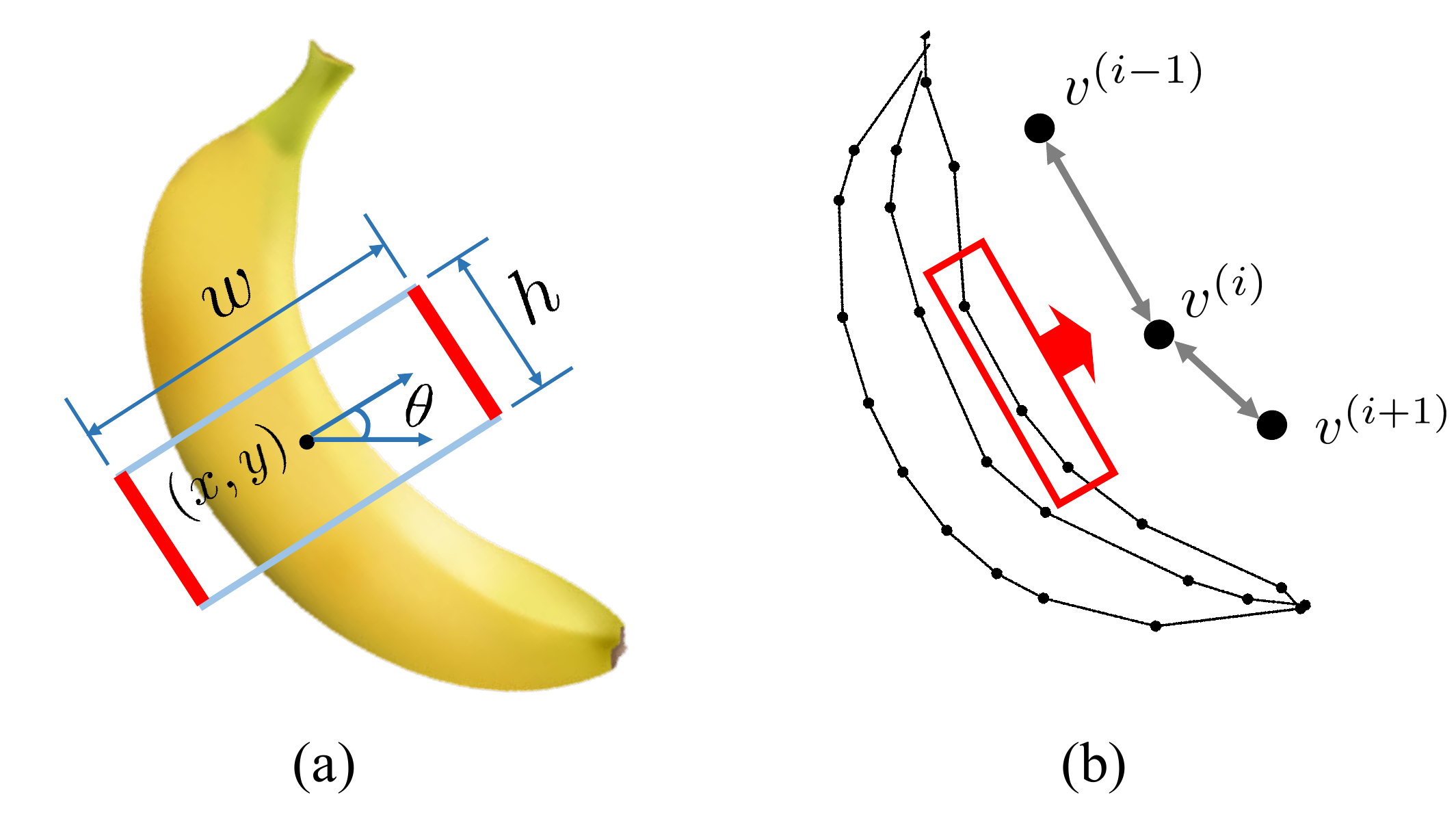}
  \vspace{-0.1in}
  \caption{(a) The grasp configuration represented by a 5-dimensional vector $(x,y,w,h,\theta)$. $(x,y)$ is the grasp center. $(w,h)$ is the width and height of grasp rectangle. $\theta$ denotes grasp orientation relative to the horizontal axis. (b) The multi-stroke sketch  represented by a directed graph $\mathcal{S}=\{\mathcal{V}, \mathcal{E}\}$. $\mathcal{V}$ and $\mathcal{E}$ indicates the vertices and edges, respectively. $v^{(i)}$ is the vertex from $\mathcal{V}$. The adjacent points $v^{(i-1)}$ and $v^{(i)}$ are connected by a directed edge (gray line). The double arrow indicates an edge in each direction. \label{fig:definition}}
  \vspace{-0.15in}
\end{figure}

\section{Problem Statement}

\noindent (a) \textbf{Grasp Configuration}: The grasp $g$ is represented by a 2D orientated rectangle, formulated as a 5-dimensional vector~\cite{lenz2015deep},
\begin{equation}
    g=\{x,y,w,h,\theta\}^T
\end{equation}
where $(x, y)$ is the grasp center, $(w, h)$ indicates the opening size and width of the end-effector of two parallel grippers, and $\theta$ denotes the grasp orientation relative to the horizontal axis as shown in Fig.~\ref{fig:definition} (a). 

\noindent (b) \textbf{RGB-D Image}: We use as input the RGB-D image, represented by a tuple:
\begin{equation}
\mathcal{I} = (\mathcal{C}, \mathcal{D})
\end{equation}
where $\mathcal{C} \in \mathbb{R}^{H'\times W' \times 3}$ is the RGB image and $\mathcal{D} \in \mathbb{R}^{H'\times W' \times 1}$ is the depth image aligned with the RGB image. The RGB image is used for searching 2D grasp rectangle, and the depth (D) image is further utilized to transfer the 2D grasp into the 3D one for robotic grasping.

\noindent (c) \textbf{Sketch}: 
\revise{The sketch is saved by recording vertices of each stroke in pixel coordinate. We further simplify these vertices on the strokes using the Ramer–Douglas–Peucker algorithm~\cite{douglas1973algorithms} into a $N_s$ simplified vertex set $\mathcal{V}=\{v^{(i)} = (v_x^{(i)}, v_y^{(i)})\}_{i=1}^{N_s}$, which are selected as the graph nodes to construct a sparse directed graph as follows,}
\begin{equation}
    \mathcal{S}=\{\mathcal{V},\mathcal{E}\}
\end{equation}
where $(v_x^{(i)}, v_y^{(i)}) \in \mathcal{V}$ denotes the 2D  coordinate of vertex $v^{(i)}$, $\mathcal{E}=\{(v^{(i)}, v^{(j)}) | (v^{(i)}, v^{(j)})\subseteq \mathcal{V}^2~and~v^{(i)}\sim v^{(j)} \}$ indicates the edges, and $\sim$ means the points of the same stroke are adjacent as shown in Fig.~\ref{fig:definition} (b). 



Take the sketch query $\mathcal{S}$ and an RGB image $\mathcal{C}$ as input, our goal is to learn a function $f$  by a parameterized neural network, which
directly outputs a grasp configuration set $g \in \mathcal{G}$ on the RGB image,
\begin{equation}
    \mathcal{G} = f(\mathcal{C},\mathcal{S})
\end{equation}
To achieve this, our network directly retrieves the potential top-k grasp configurations $\mathcal{G}=\{g_1, g_2, \cdots, g_k\}$ upon  target depicted objects, conditioned on  sketch query $\mathcal{S}$.

\begin{figure*}[thpb]
  \centering
  \includegraphics[width=0.9\textwidth]{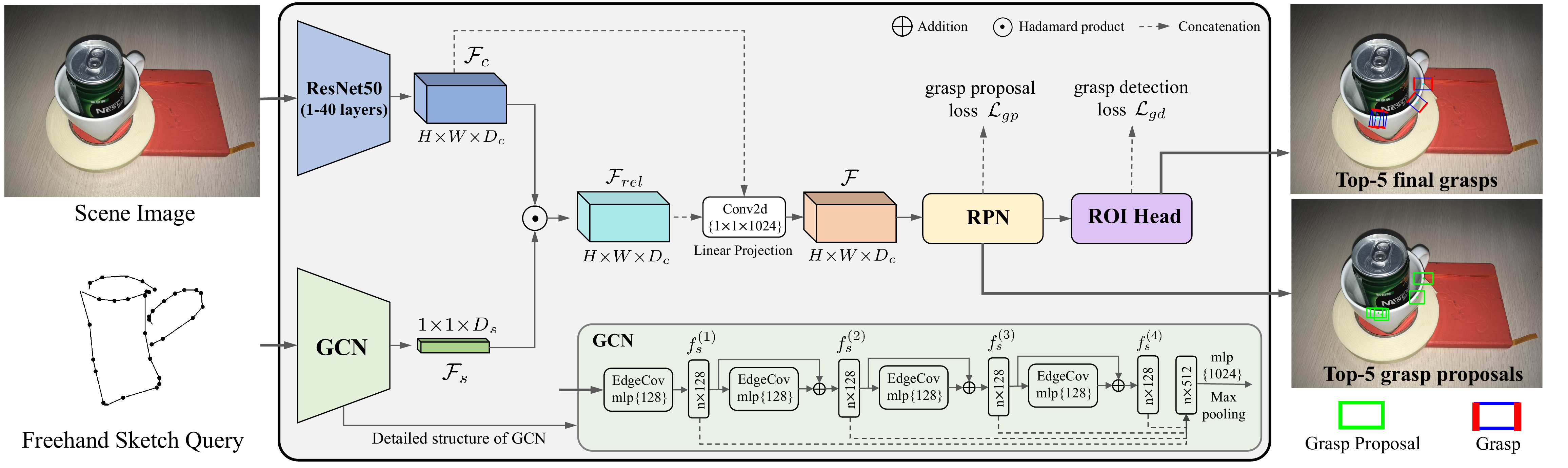}
  \vspace{-0.1in}
  \caption{Architecture of our sketch-conditioned grasp detection network. The network takes as input a natural scene image and a graph-represented sketch. The relevant maps $\mathcal{F}_{rel}$ are calculated by the extracted image features $\mathcal{F}_c$ and sketch features  $\mathcal{F}_s$. Finally, the feature maps $\mathcal{F}$ are fed into the subsequent RPN for learning conditional grasp proposals. The final grasps represented by a 5-dimensional vector are predicted from the ROI head.
  \label{fig:architecture}}
  \vspace{-0.15in}
\end{figure*}

\section{Method}
\subsection{Feature Extraction}
This step aims to extract the useful feature representation from target natural RGB images $\mathcal{C}$ and sketch queries $\mathcal{S}$. For the RGB image, the network demands to extract the geometric information relevant to the graspable regions on the objects and semantic information relevant to the attribute of the object. We use the ResNet-50~\cite{he2016deep} pre-trained on the ImageNet as an encoder $\Phi_c$ to extract the feature representation $\mathcal{F}_c \in \mathbb{R}^{H \times W \times D_c}$ of the natural image,
\begin{equation}
    \mathcal{F}_c=\Phi_c(\mathcal{C})
\end{equation}

A vanilla way is to take the sketch query as an image and then extract the feature representation $\mathcal{F}_s$ by using the ResNet, making an analogy to feature extraction $\mathcal{F}_c$ of natural images aforementioned.  However, this operation ignores the stroke structure and relation of adjacent points of the stroke. Therefore, to utilize the structure information of the sketch, we build the sketch query as a graph. Inspired by the graph neural network utilized in the sketch segmentation task, we adopt a similar structure as~\cite{yang2021sketchgnn}. The Graph Convolutional Network (GCN) is composed of several modules of EdgeConv~\cite{wang2019dynamic} connected in a residual manner, as shown in Fig.~\ref{fig:architecture}. The output features after each EdgeConv module are represented as $\{f_s^{(i)}\}_{i=1}^{4} \in \mathbb{R}^{N_s \times 128}$, where $N_s$ is the vertex number of the sketch. Then we concatenate features $\{f_s^{(i)}\}_{i=1}^{4}$ extracted from low and high levels, and aggregate them by max pooling operation to obtain a global feature vector $\mathcal{F}_s \in \mathbb{R}^{1\times 1\times D_s}$. We denote the graph-based encoder as $\Phi_s$ and summarize the feature extraction from the graph-represented sketch as,
\begin{equation}
    \mathcal{F}_s=\Phi_s(\mathcal{S})
\end{equation}

\subsection{Feature Querying}
In this stage, we utilize the global feature representation $\mathcal{F}_s$ to query the potential location of the spatial feature maps $\mathcal{F}_c$. Concretely, we perform Hadamard product to calculate the relevance between the spatial feature maps against the global representation $\mathcal{F}_s$, \ie, $\mathcal{F}_{rel} = \{\mathcal{F}_c^{(i)} \odot  \mathcal{F}_s\}_{i=1}^{H\times W}$, where $\mathcal{F}_c^{(i)} \in \mathbb{R}^{1\times 1 \times D_c}$ denotes the each location of the spatial feature maps $\mathcal{F}_c$. 
This mechanism enhances the response of features highly relevant to the sketch query, otherwise suppressing that of irrelevant locations on the feature maps. It also means that we use the global sketch feature $\mathcal{F}_s$ to query the corresponding locations of the target image feature maps $\mathcal{F}_c$. We concatenate the original spatial feature maps $\mathcal{F}_c$ and the relevance maps $\mathcal{F}_{rel}$, on which we then perform linear projection to get the final feature maps $\mathcal{F} \in \mathbb{R}^{H \times W \times D_c}$.

\subsection{Region Proposal and Classification}
Finally, we send final feature maps $\mathcal{F}$ to the region proposal network (RPN) to retrieve grasp proposals relevant to the sketch query. Concretely, the feature maps $\mathcal{F}$ are fed into two sibling fully-connected layers which respectively predicts bounding box $\hat{t}^{(i)} = (\hat{x}, \hat{y}, \hat{w}, \hat{h})$ and score $\hat{p}^{(i)}$ of grasp proposal from each anchor $i$. That is, we predict a matching score for each grasp proposal and filter the sketch-irrelevant grasp proposals with a lower confidence score. 
Then we treat the grasp orientation as a classification task, which discretizes the grasp orientation $\theta$ as $\{10^{\circ}, 20^{\circ}, \cdots, 180^{\circ}\}$, totaling $N_{ori}=18$ classification labels $\{{c^{(i)}=i}\}_{i=1}^{N_{ori}} $ for orientation quantification. 
The additional classification label $c^{(0)}=0$ indicates that the grasp candidate is not relevant to the sketch query.
Therefore, the total number of classes is 19. Specially, this stage classifies the queried grasp proposals for potential orientation $\theta$, and refines the current grasp proposal box to approximate the target bounding box $(x,y,w,h)$.

\subsection{Loss Function} To efficiently train our model, two separate losses are used: grasp proposal loss $
\mathcal{L}_{p}$ and grasp detection loss $
\mathcal{L}_{g}$.

The grasp proposal loss $
\mathcal{L}_{gp}$ is utilized to supervise the RPN branch, which is similar to~\cite{ren2015faster}.
\begin{equation}
\begin{aligned}
    \mathcal{L}_{gp}(\{\hat{p}^{(i)}\}, \{\hat{t}^{(i)}\}) = &\frac{1}{N_{cls}} \sum_{i} L_{cls}(p^{(i)},\hat{p}^{(i)})\\
    &+ \frac{1}{N_{reg}} \sum_{i} {p}^{(i)} L_{reg}(t^{(i)},\hat{t}^{(i)})
\end{aligned}
\end{equation}
where $\hat{p}^{(i)}$ is the predicted probability that the anchor $i$ in a mini-batch belongs to the target sketch-queried proposal; 
$t^{(i)}$ and $\hat{t}^{(i)}$ are the ground-truth and predicted grasp proposal regions represented by 4-dimensional vectors $(x,y,w,h)$ and $(\hat{x}, \hat{y}, \hat{w}, \hat{h})$, respectively. 
We assign $p^{(i)}=0$ for no grasp relevant to the sketch query otherwise $p^{(i)}=1$.
 $N_{cls}$ is the mini-batch size of 256 during the RPN training stage, and ${N_{reg}}$ is set as 256. $L_{cls}$ is the cross-entropy loss of grasp proposal classification (target or non-target grasp proposals), and $L_{reg}$ is the smooth loss of grasp proposal regions.


Supervised by the grasp proposal loss $\mathcal{L}_{gp}$, the network learns to predict the grasp proposals $\hat{t}^{(i)}$ matching the corresponding sketch query. For each positive region proposal $
\hat{t}^{(i)}$, the ROI head needs to regress the grasp region $(x,y,w,h)$ and classify the orientation $\theta$ of grasp rectangle. The grasp detection loss is defined as follows,
\begin{equation}
\begin{aligned}
    \mathcal{L}_{gd}(\{\hat{c}^{(j)}\}, &\{\hat{g}^{(j)}\}) = \frac{1}{N_{cls}^g} \sum_{j} L_{cls}(c^{(j)},\hat{c}^{(j)}) \\+ & \frac{1}{N_{reg}^g}\sum_{j}\bm{1}(c^{(j)} \neq 0)L_{reg}(t^{(j)},\hat{g}^{(j)}_{0:3})
\end{aligned}
\end{equation}
where $c^{(j)}$ and $\hat{c}^{(j)}$ are the ground-truth and predicted class of the sampled grasp proposal $j$; $N_{cls}^g=N_{reg}^g=512$ are the number of sampled grasp proposals. 
The indicator function $\bm{1}(\cdot)$  is assigned to 1 for positive (target) grasp proposals ($c^{(j)}\neq 0$) otherwise 0. 

Finally, the model is trained in an end-to-end manner by the joint training loss:
\begin{equation}
    \mathcal{L}=\mathcal{L}_{gp}+\mathcal{L}_{gd}
\end{equation}

The joint training of these multiple tasks enables our model to reduce accumulative errors between each task, as the whole network jointly optimizes and balances the errors of the individual task to enhance the overall performance. 

\section{Implementation Details}
\noindent \textbf{Network Architecture.} As shown in Fig.~\ref{fig:architecture}, we use the ResNet-50 (1-40 layers) pre-trained on the ImageNet to extract the image features, and the image feature maps $\mathcal{F}_c$ have the dimension of $D_c = 1024$. For the sketch, we use the Graph Neural Network to extract the global features $\mathcal{F}_s$ with the dimension of  $D_s = 1024$. 

\noindent \textbf{Dataset.} We evaluate the performance of our method on three different datasets, including VMRD, GraspNet-1Billion (G-1Billion), and our collection of household objects, along with a sketch dataset QuickDraw. 

\noindent (1) \textit{VMRD}~\cite{zhang2018visual}. The original VMRD dataset contains 4683 images of clutter scenes, along with annotated grasps upon 17000 objects from 31 categories. The dataset is split into a training set of 4233 images and a testing set of 450 images. There are a total of 12  categories both existed between QuickDraw and VMRD datasets. 

\noindent (2) \textit{G-1Billion}~\cite{fang2020graspnet}. The G-1Billion dataset contains densely annotated grasp rectangles of multiple object scenes. To cover more object categories, we use the test dataset of G-1Billion, where contains 90 scenes and 88 objects. The images in each scene are captured from 256 different dense views. We sample images of each scene by 5 equal-length intervals to obtain 52 images each scene. 
Then the first 42 images are split as training set and the last 10 images as the testing set, totaling 3780 training images and 900 testing images, respectively. 
Therefore, the split dataset demands our algorithm robust against novel diverse sketches and camera poses unseen during the training stage. 
We utilize the Non-Maximum Suppression (NMS) algorithm to filter the boxes with an exceeded ratio overlap of 0.7 from the dense 2D grasp labels. Finally, we choose the common 20 categories between QuickDraw and G-1Billion datasets. 

\begin{figure}[thpb]
  \centering
  \includegraphics[scale=0.3]{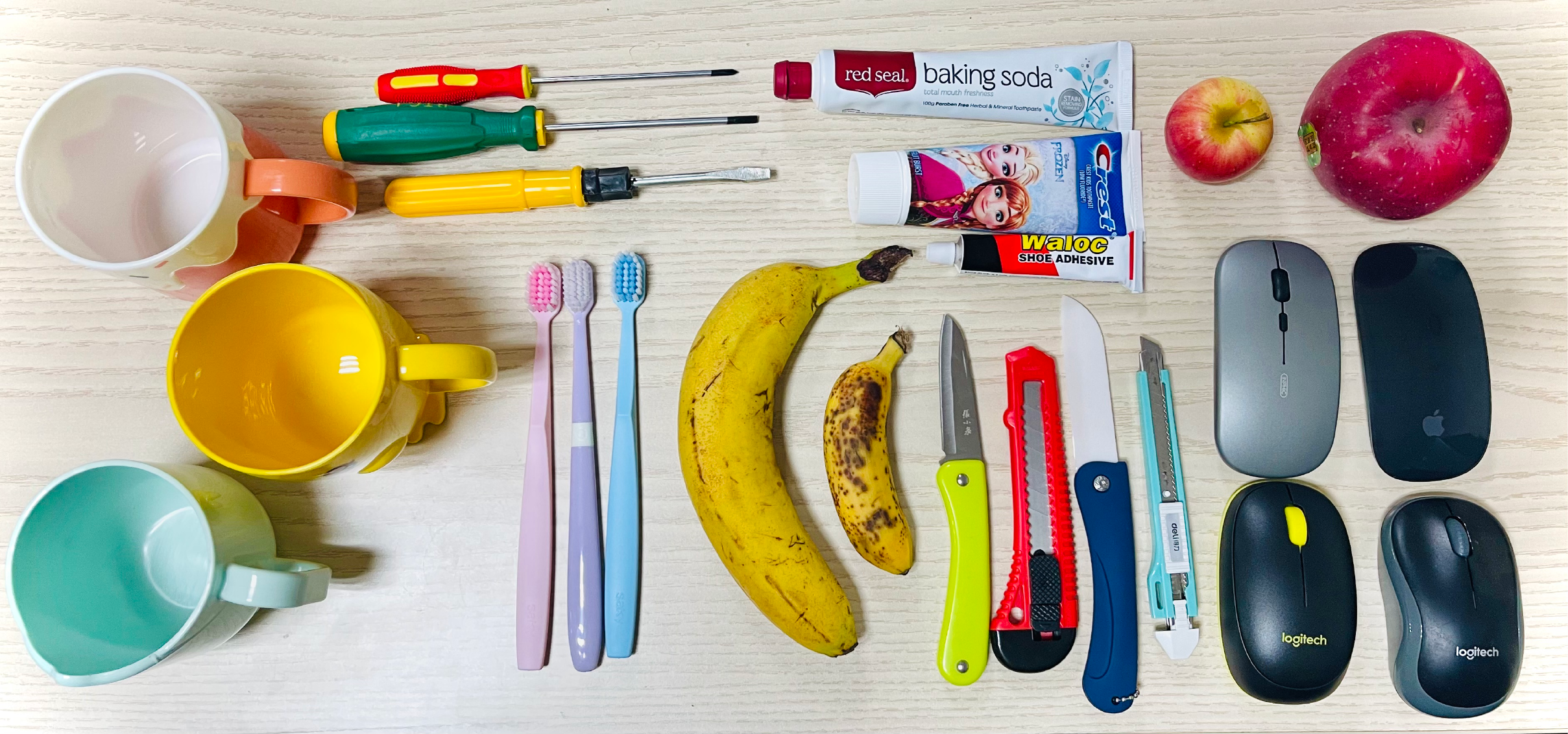}
  \vspace{-0.1in}
  \caption{The 24 household objects are used for physical robot experiments. The testing object collection contains instances from 8 categories, including \textit{apple}, \textit{banana}, \textit{cup}, \textit{knife}, \textit{mouse}, \textit{screwdriver}, \textit{toothbrush} and \textit{toothpaste}.\label{fig:objects}}
  \vspace{-0.15in}
\end{figure}

\noindent (3) \textit{Our object collection}. To validate the generalization of our method, we collect 24 household objects with different shapes and textures compared to the objects in the training data. The collection of objects is shown in Fig.~\ref{fig:objects}.

\noindent (4) \textit{QuickDraw}~\cite{jongejan2016quick}.
We randomly choose 12000 sketches for each category, which are  split into 9600 and 2400 for training and testing, respectively. For the testing set (2400 images), we split it into two equal part: testA (1-1200 images) and testB (1200-2400 images). During testing, we evaluate the performance of our model on the two sketch testing set (testA and testB), to further validate the generalization of model on different unseen sketches.

\noindent \textbf{Training and testing.} Our network is implemented on PyTorch. We train our network on a single RTX 2080Ti GPU with a batch size of 8. The SGD optimizer with a monument of 0.9 and a weight decay of 0.0005 is used. We set the learning rate as 0.005, which is decayed by a factor of 0.75 for every 2600 iterations. We train the model for nearly 50K iterations. 
During the training, we apply data augment, including color jittering, brightness enhancement, and random flip to avoid over-fitting. Specially, the scene backgrounds of G-1Billion dataset are relatively fixed. To make our model deal with different scenes, we randomly choose an image from the SUN2012~\cite{xiao2010sun} dataset to replace the original background by utilizing the provided object masks of G-1Billion.
During the testing, the top 300 proposals are used for grasp prediction. The NMS algorithm is applied to preserve the final grasps.

\section{Experiments}
\subsection{Evaluation Metrics.}
\noindent \textbf{Grasp Detection on Benchmarks.} We adopt  top-k precision (P@k) and recall (R@k) metrics as in~\cite{chen2021joint} to evaluate sketch-conditioned grasp detection on two public benchmarks. P@k is the proportion of correct grasps from retrieved top-k grasps, and R@k calculates the proportion of at least one correct grasp from the top-k grasps. The detected grasp is considered as a correct one if it simultaneously satisfies following conditions: the predicted grasp $\hat{g}$ has a Jaccard Index~\cite{zhang2019roi} larger than $0.25$ and an angle error less than $30^\circ$ with at least one of the target ground truth grasps $g$. 

\noindent \textbf{Physical Robot Experiments.}
In the physical experiments, we respectively calculate the success rate of grasp detection and grasp execution. The model is trained on the VMRD and G-1Billion datasets. The grasp detection is considered correct if it is visually located in reasonable graspable regions. For the grasp execution, the successful grasp means that the robot follows the predicted grasp rectangle to grasp and place the sketch-depicted object out of the scene. We try 15 attempts for each category and record the total number of success. 

\subsection{Comparison with Baseline Methods}
\noindent \textbf{Baseline Methods.}
To validate the efficacy of our proposed method, we list some baseline methods as follows,

\noindent (1) \textit{Random-grasp}: We use the category-agnostic multi-object grasp detection method~\cite{chu2018real} as a baseline and retrain the network on VMRD and G-1Billion dataset individually. 

\noindent (2) \textit{Object\&Grasp}: Similar to the work of~\cite{chen2021joint}, we incorporate two networks in a cascaded manner to generate the final grasp.
We first use the state-of-the-art sketch-guided object localization network~\cite{tripathi2020sketch} to detect target object's box and then use grasp detection network~\cite{chu2018real} to search all grasp candidates within the detected object box. The grasp candidate is selected as the final grasp if it has the minimum distance to the center of detected object box. Both models are retrained.

\noindent (3) \textit{Ours(image)}: This is a variant of our model, which takes the sketch image as input but not a graph-represented one. The GCN is therefore replaced with the ResNet-50 to fit the input of the sketch images.

\begin{figure*}[thpb]
  \centering
  \includegraphics[width=0.9\textwidth]{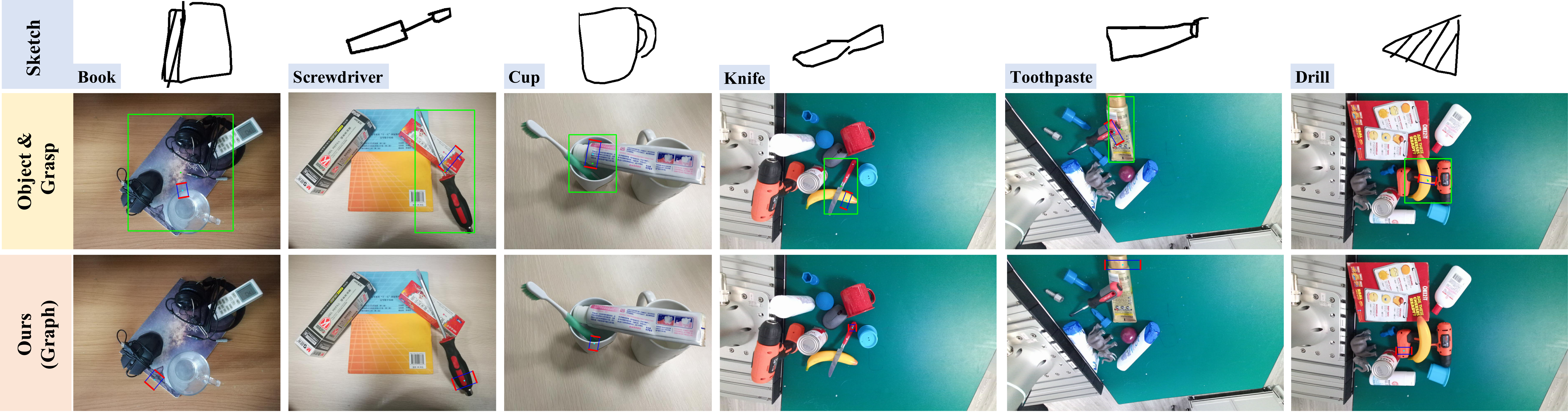}
  \vspace{-0.1in}
  \caption{Grasp detection results for \textit{Object\&Grasp} and \textit{Ours(graph)} on VMRD (first 3 columns) and G-1Billion (last 3 columns) datasets. The green box is the object-level detected results, and the red box is the grasp-level detected oriented rectangle. The red grasp box within the green object box is assigned according to the principle of the minimum distance between the grasp center and object center. \label{fig:compare_vis}}
  \vspace{-0.15in}
\end{figure*}

We report the top-k precision and recall results in Tab.~\ref{tab:compare_dataset}. The method of \textit{Random-grasp} has lower performance than the other three methods on two benchmarks, especially at the precision metric. Compared to the cascaded method \textit{Object\&Grasp}, two of our methods outperform it significantly. Additionally, we show the weakness of the cascaded method by exemplars in Fig.~\ref{fig:compare_vis}. In the cluttered scene, the cascaded method is easily confused by the non-target objects within the detected object box, and even if the object box is localized precisely, it can lead to failed cases. 
\revise{In contrast, direct conditional grasp detection is less influenced by overlapping non-target objects, as it usually detects small grasp regions of a single object, rather than matches one of grasp regions within a large detected object region.
}
Our graph-based model has comparable results against the image-based model at the recall metric and outperforms the image-based model by nearly 2\% in terms of precision. 
As an image-based model uses the ResNet50, which should have a larger capacity versus the graph-based model of GCN (parameter size of 24.18M vs. 0.63M), the graph-based model still overall slightly outperforms the image-based model. \revise{Compared to the cascaded pipeline \textit{Object\&Grasp}, our end-to-end pipeline is also more efficient given an input scene image with a resolution of 1280 $\times$ 720 (runtime of 750 ms vs. 430 ms).}

\begin{table*}[]
\centering
\renewcommand\tabcolsep{2.5pt}
\renewcommand{\arraystretch}{1.0}
\footnotesize
\caption{Comparison results for different methods tested with different groups of unseen sketches testA/testB. \label{tab:compare_dataset}}
 \vspace{-0.1in}
\begin{tabular}{c|c|cccc|cccc}
\toprule[1pt]
\multirow{2}{*}{Dataset} & \multirow{2}{*}{Method} & \multicolumn{4}{c|}{Top-k Recall} & \multicolumn{4}{c}{Top-k Precision} \\
                            &         & R@1 & R@3 & R@5 & R@10 & P@1   & P@3 & P@5 & P@10 \\ \midrule[0.5pt]
\multirow{4}{*}{VMRD~\cite{zhang2018visual}}       & Random-grasp & 27.73 & 54.78 & 70.26 & 83.71 & 27.73 & 27.41 & 28.37 & 27.43 \\
                            & Object\&Grasp &  50.54/51.01 & 59.30/60.05 & 64.42/66.13 & 72.91/74.36 & 50.54/51.01 &51.43/52.68 & 51.83/53.44 & 51.91/53.41 \\
                            & Ours(image)  &70.12/70.26 & 85.33/\textbf{86.27} & \textbf{89.37}/90.04 & 92.33/92.60& 70.12/70.26  & 69.58/70.21& 68.66/69.51 & 67.80/68.79 \\
                            & Ours(graph) & \textbf{71.47}/\textbf{70.79} &\textbf{85.73}/85.73 &88.56/\textbf{90.58}& \textbf{94.21}/\textbf{94.21}& \textbf{71.47}/\textbf{70.70} & \textbf{71.62}/\textbf{70.30} & \textbf{70.59}/\textbf{69.93} & \textbf{69.36}/\textbf{69.41} \\ \midrule[0.5pt]
\multirow{4}{*}{G-1Billion~\cite{fang2020graspnet}} & Random-grasp & 9.79  &25.83 & 35.36 & 53.48 & 9.79 & 10.27 & 10.04 & 10.08 \\
                            & Object\&Grasp & 51.13/51.80 & 57.44/57.85 & 60.63/60.74 & 65.27/65.08 & 51.13/51.80 & 49.53/50.27 & 48.22/48.71 & 43.82/43.99 \\
                            & Ours(image)  &  55.00/54.56   & \textbf{65.27}/64.38 & \textbf{69.83}/69.09 & \textbf{76.09}/\textbf{75.69} & 55.00/54.56 & 55.03/54.32 & 55.27/54.23 & 54.90/54.31  \\
                            & Ours(graph) & \textbf{56.75}/\textbf{56.60} & 64.98/\textbf{65.64} & 69.16/\textbf{69.26} & 74.80/73.80 & \textbf{56.75}/\textbf{56.60} & \textbf{56.65}/\textbf{57.01} & \textbf{56.59}/\textbf{56.65} & \textbf{56.25}/\textbf{55.90} \\ \bottomrule[1pt]
\end{tabular}

 \vspace{-0.15in}
\end{table*}

\begin{figure}[t]
  \centering
  \includegraphics[scale=0.4]{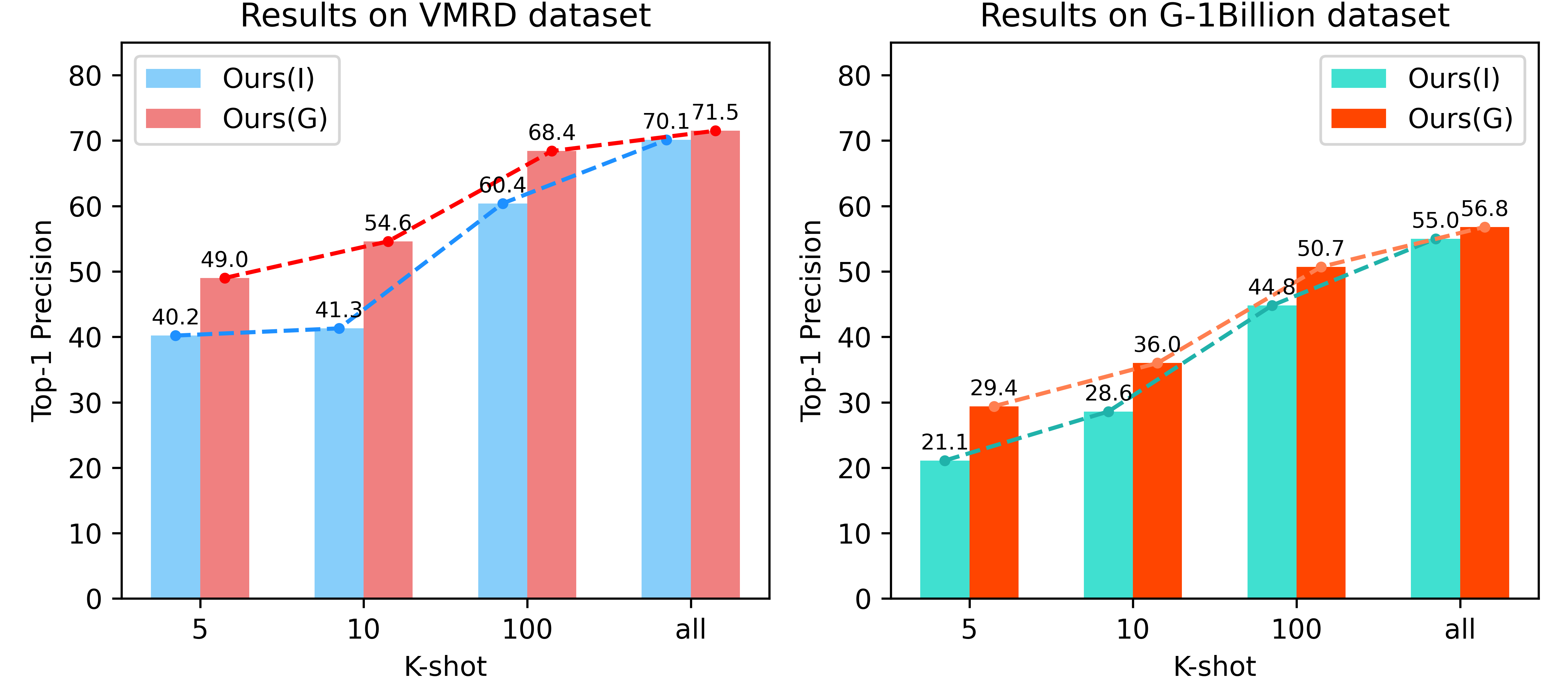}
    \vspace{-0.1in}
  \caption{Top-1 precision results under different \textit{K-shot} settings by using unseen sketch testing set (TestA). Our image-based model Ours(I) and graph-based model Ours(G) are trained with 5, 10, 100, and all training samples of sketch, respectively. \label{fig:few-shot}}
  \vspace{-0.15in}
\end{figure}

\noindent \textbf{Few-Shot task performance.}
Due to limited training samples of sketches in practical application, we also conduct the experiments under the few-shot setting to validate the efficacy of our graph-based model. Concretely, we only choose 5, 10, and 100 samples to train our image-based and graph-based model and report the results in Fig.~\ref{fig:few-shot}.
Our graph-based model Ours(G) significantly outperforms the image-based model Ours(I) by a large margin of nearly 8\% in the 5-shot setting, as the graph characterizes the sketch structure, which facilitates the network to extract more helpful feature representation. The experiment indicates the capability of our graph-based model in real-world applications, as it only uses fewer model parameters and training samples than that of image-based one. Extending our graph-based model to be adaptive with more categories is feasible by collecting a handful of sketch data to train the network, avoiding the tedious collection of extensive training data.

\noindent \textbf{Impact of graph node density.} \revise{We test the same model used in Tab.~\ref{tab:compare_dataset} on the VMRD dataset, however, where the nodes of sketch are up-sampled and sub-sampled into $2N_s$ and $N_s/2$, respectively. The top-3 precision results under these two densities are 67.03\% (up-sampled) and 63.19\% (sub-sampled), which drop by a margin compared to that in Tab.~\ref{tab:compare_dataset}. Thus, missing sketch details affect results more than increasing details. These results validate that our model is still robust against different densities of graph nodes.
}

\subsection{Real Robot Experiments}
\noindent \textbf{Hardware.} We conduct physical experiments on the Rethink Baxter robot with an Intel RealSense D435 camera overlooking the table. The Baxter robot has two 7-DoF arms with grippers. 
Our network and control algorithm are implemented on a desktop computer with an Intel i5-9400F CPU and a single NVIDIA RTX 2060S GPU. As our model predicts the grasp configuration on the RGB image without scale information, the depth (D) information is utilized in practical grasping. We adopt the grasp strategy mentioned in~\cite{lenz2015deep}.


\noindent \textbf{Isolated unseen object scenes with unseen sketches.}
We choose instances from the 8 categories in Fig.~\ref{fig:objects} and place them on the table with arbitrary poses. We randomly choose a single instance with the sketch matching the chosen object each time, totaling tested with 10 trials for each category. Moreover, another 5 trials are when the presented object is not matched with the given sketch query. The sketches used in the experiments are all unseen in training and randomly chosen from the Quickdraw. We report the results in Tab.~\ref{tab:robot_single}.

\begin{table}[]
\centering
\renewcommand\tabcolsep{3pt}
\renewcommand{\arraystretch}{0.95}
\footnotesize
\caption{Results on scenes with unseen single objects.\label{tab:robot_single} }
  \vspace{-0.1in}
\begin{tabular}{c|ccc|cc}
\toprule[1pt]
\multirow{2}{*}{Objects} &
  \multicolumn{3}{c|}{Object\&Grasp} &
  \multicolumn{2}{c}{Ours} \\
              & Object  & Grasp & Execute & Grasp & Execute \\ \midrule[0.5pt]
Apple         & 14/15 & 14/15  & 14/15    & 15/15 & 14/15   \\
Banana        & 14/15 & 14/15  & 14/15    & 14/15  & 14/15    \\
Cup           & 10/15 & 10/15  & 10/15    & 12/15  & 11/15    \\
Knife         & 11/15 & 11/15  & 11/15    & 12/15  & 12/15    \\
Mouse         & 10/15 & 10/15  & 10/15    & 10/15  & 10/15    \\
Screwdriver   & 10/15 & 10/15  & 10/15    & 11/15  & 11/15    \\
Toothbrush    & 11/15 & 11/15  & 9/15    & 12/15  & 12/15    \\
Toothpaste    & 11/15 & 11/15  & 11/15    & 12/15  & 12/15    \\ \midrule[0.5pt]
{Average(\%)} & {75.83} & {75.83} & {74.17} & {81.67} & {80.00} \\ \bottomrule[1pt]
\end{tabular}
 \vspace{-0.15in}
\end{table}

\noindent \textbf{Multiple unseen object scenes with unseen sketches.}
We arbitrarily choose a single instance with a matching sketch for each category, and other 3-5 irrelevant objects are randomly selected from the remaining 7 categories. The selected objects are placed on the table, and there are varying levels of overlaps among objects. It is more challenging than single-object scenes due to (1) large variations between unseen testing objects and training data, and (2) similarities between testing sketches (\eg, \textit{screwdriver} vs. \textit{toothbrush}).
Each category is tested with 15 trials. The experiment results are summarized in Tab.~\ref{tab:robot_multiple}. The results show that our end-to-end method greatly outperforms the cascaded one in multi-object scenes, as the cascaded method will be easily influenced if the non-target object is close to the center of the target object (\eg, a \textit{toothpaste} on the top of a \textit{cup}), resulting in poor performance.

\begin{table}[]
\centering
\renewcommand\tabcolsep{3pt}
\renewcommand{\arraystretch}{0.95}
\footnotesize
\caption{Results on scenes with unseen multiple objects.\label{tab:robot_multiple}}
 \vspace{-0.1in}
\begin{tabular}{c|ccc|cc}
\toprule[1pt]
\multirow{2}{*}{Objects}  & \multicolumn{3}{c|}{Object\&Grasp} & \multicolumn{2}{c}{Ours} \\
              & Object & Grasp & Execute & Grasp & Execute \\ \midrule[0.5pt]
Apple         & 14/15  & 13/15 & 13/15   & 14/15 & 14/15    \\
Banana        & 13/15   & 10/15  & 10/15    & 12/15  & 12/15    \\
Cup           & 10/15   & 7/15  & 5/15    & 11/15  & 9/15    \\
Knife         & 7/15   & 7/15  & 6/15    & 9/15  & 8/15    \\
Mouse         & 8/15   & 8/15  & 8/15    & 10/15  & 10/15    \\
Screwdriver   & 8/15   & 6/15  & 5/15    & 8/15  & 7/15    \\
Toothbrush    & 5/15   & 5/15  & 5/15    & 8/15  & 7/15    \\
Toothpaste    & 10/15   & 8/15  & 7/15    & 10/15  & 10/15    \\ \midrule[0.5pt]
{Average(\%)} & 62.50     & 53.33     & 49.17     & 68.33       & 64.17       \\ \bottomrule[1pt]
\end{tabular}
 \vspace{-0.15in}
\end{table}

\section{CONCLUSIONS}
This paper proposes an end-to-end conditional grasp detection network associated with the drawing sketch. We build the sketch as a graph by taking the structure of the sketch into account, facilitating the network to extract more representative features to differentiate the sketches. Moreover, this graph representation enhances the network's generalization ability by using a handful of sketch samples. We achieve state-of-the-art performance on the VMRD and G-1Billion benchmarks. The implementation of our network on the physical Baxter robot shows the capability of our model on clutter scenes with multiple objects.










\clearpage
\bibliography{./root}      
\bibliographystyle{./IEEEtranS.bst}

\end{document}